# Towards reducing the multidimensionality of OLAP cubes using the Evolutionary Algorithms and Factor Analysis Methods.

Sami NAOUALI[1] and Semeh BEN SALEM [2]

[1]Virtual Reality and Information Technologies LAB, Military Academy of Fondouk Jedid, TUNISA
`snaouali@gmail.com`
[2] Virtual Reality and Information Technologies LAB, Military Academy of Fondouk Jedid, TUNISA
`Semeh.bensalem@yahoo.fr`

*Abstract*

*Data Warehouses are structures with large amount of data collected from heterogeneous sources to be used in a decision support system. Data Warehouses analysis identifies hidden patterns initially unexpected which analysis requires great memory and computation cost. Data reduction methods were proposed to make this analysis easier. In this paper, we present a hybrid approach based on Genetic Algorithms (GA) as Evolutionary Algorithms and the Multiple Correspondence Analysis (MCA) as Analysis Factor Methods to conduct this reduction. Our approach identifies reduced subset of dimensions p' from the initial subset p where p'<p where it is proposed to find the profile fact that is the closest to reference. GAs identify the possible subsets and the Khi² formula of the ACM evaluates the quality of each subset. The study is based on a distance measurement between the reference and n facts profile extracted from the warehouse.*

*Keywords*

 Data Warehouse,   OLAP cube,   dimensions reduction,   Genetic Algorithms,   Multiple Correspondence Analysis,

## 1. Introduction.

Data Warehouses (DW) are used to collect and store great data quantity in large tables to be then processed which helps extracting knowledge and assist in decision making. DW collects the data stored from heterogeneous sources and creates centralized huge data bases to identify initially hidden patterns once treated. In (W. INMON, 1994), the authors present these large data structures and their architecture as a result of the data gathering from centralized relational databases initially decentralized which can homogenize them and offer the possibility to treat and process them which is initially a hard task. Graphically, these data sets can be represented by cubic multidimensional data structures called OLAP cubes. (L.Lebart, A.Morineau, M.Piron, 2006) consider the multidimensional statics as an interesting analysis axis that should be studied. The random assembly of data and their arrangement in the DW may contain unnecessary and superfluous data that have no real interest for the analyst and can even provide incorrect results if included in the processing. To facilitate the exploitation and interpretation of the DW, it is necessary to propose innovative methods in order to reduce the huge amount of data contained in the DW. The reduction methods proposed focuses mainly on reducing either the lines or the columns of the DW.

In this context, we propose in this paper an approach based on the use of two methods: Genetic Algorithms (GAs) and the Multiple Correspondence Analysis (MCA). The combination of these two methods makes it possible to converge, using a specified process, to an optimum solution in a set of possible ones. This new solution that has been identified helps identifying the most appropriate dimensions that can be reduced and determining the best subset of dimensions that best summarizes the DW.

In this paper, the second section details the previous work in relation with the topic of reduction and exposes their limitations. The third section presents the two techniques we suggest to use in our approach. The third section explains the approach and the adaptation of the tools used to the multidimensional context, an algorithm representing the whole process is also presented in this section. In section four, we present an illustrative example and we finish by interpreting the results and giving our conclusion and perspectives.

## 2. Related Works and Motivations

The objective of reducing the data in a DW is to provide greater clarity of the information stored and facilitates their processing which keeps only the information with a real interest. In this context, several proposals were made, either by removing data stored for a long period of time, aggregating them or reduce the dimensions and facts. Each of these approaches has its limits we propose to identify in the following paragraph.

In (S.Naouali, 2012), the author proposes a hybrid approach to reduce the dimensions of a DW by applying the Rough Sets Theory (RST)(Pawlak, 1995) and the Principal Component Analysis (PCA) (I. Jolliffe, 2002). The RST identifies and removes redundant and needless facts while the PCA is used for quantitative data tables to reduce the number of dimensions. However, the reduction represents in its self a loss of information seeing that some dimensions will be ignored. On the other hand, this approach concerns only great quantitative data tables.

In (J. Skyt, C.S. Jensen, T.B. Pedersen, 2001),the authors propose to proceed by aggregating the data to move to a higher level of granularity of the DW dimensions.This technique can lead to the loss or non-visibility of some details which is binding for the analyst. The use of aggregation functions allows passing from a granularity level to another. Such a methodology can hide certain details with big importance for the analyst and as a result can bring confusion.

(C. S. Jensen, 1995) proposes to remove the historical data that has exceeded a certain limit, considering that such information is no longer useful given the time elapsed which can be critical because the deleted data may contain useful information and can power a recent model using old stored data.

(W.Wang, J.Feng, H.Lu, J.Xu Yu, 2002) propose an approach towards reducing the lines (facts) of a cube by creating a condensed cube with a smaller size than the non condensed one which would help reduce its computation time.The approach explores the properties of single tuples so that a number of tuples in a cube can be condensed into one without loss of any information.

In our approach, we propose to use one of the methods of factor analysis, the MCA associated with the theory of Genetic Algorithms and consider the problem as an approximation approach: we identify first the possible subsets in concordance with the problematic and then evaluate the quality of each solution to either retain it or remove it.

## 3. Required Prior Knowledge and Techniques.

### 3.1 Evolutionary Algorithms.

Evolutionary algorithms are stochastic algorithms inspired from the theory of evolution of species, and used to solve optimization problems. They define a set of possible solutions evaluated to converge to the best results. Three major families of algorithms have been proposed as evolutionary algorithms:

The evolutionary strategies (H.-G. Beyer, H.-P. Schwefel,2002), to solve continuous optimization problems.

Evolutionary Programming, considered as a method of artificial intelligence to the finite state automata design.

Genetic algorithms (D. Goldberg, 1994).

In our approach, we use the GAs as optimization algorithms which process helps identifying the most optimal solution to a specified problem. Each individual is represented by a chromosome that represents a potential solution and must meet the constraints of the problem enunciated. The initial population contains N individuals randomly coded. A fitness function will be used to evaluate the quality of each solution contained in a population. A selection operator (J .Baker, 1995) will be then applied to keep individuals who were considered sufficiently relevant and can give birth to more optimal solutions. The most optimal solutions obtained will be then included in the next generation i+1 using the selection operator. The multiplication of the individuals is the same generation is executed with the crossover operator and the mutation operator.

### 3.2 Factor Analysis Methods.

Factor analysis is a descriptive method used to describe a set of observed variables by latent ones (not observed). This method of analysis helps reducing the data amount in great data tables by reducing the number of variables. Each factorial method is associated to a particular type of data. The most known factor analysis methods are:

The Principal Component Analysis (PCA) for quantitative variables.
The Multiple Correspondence Analysis (MCA) for qualitative variables.
Mixed Data Factor Analysis (MDFA) for mixed tables.
The Multiple Factor Analysis (MFA) for mixed tables grouped data.
The Hierarchical Multiple Factor Analysis (HMFA) to generalize the AFM.
The Factor Analysis of Correspondences (FAC) is dedicated to contingency tables.

In our approach, we propose to use, in the process of the GAs, a fitness function extracted from the MCA. The choice of the fitness fitness is very important in this study. It should meet the constraint of the approach to avoid producing mistaken results. The fitness fucntion we propose to exploit corresponds to the Khi² formula used to compute the similarity between two individuals by measuring the distance betwwen their respective caracteristics. The MCA is used for large qualitative data tables. A data table with I individuals and J variables is then represented by a matrix with I rows and J columns, $x_{ij}$ represents the modality of the variable j possessed by the individual i. In (J.Pagès, F.Husson, 2009), the authors present special tables of the MCA: Table of Condensed Coding (TCC), specified by the user and submitted to the processing software for analysis. The Complete Disjunctive Table (CDT) that contains in lines the individuals as presented in the TCC, and in columns the different modalities. If $k_j$ is the number of modalities of the variable j, the total number of modalities is then $K = \sum_j k_j$. The number of individuals in the rows corresponds to the same number as in the CDT. The total number of modality is $K_j$ and we have:

$$\begin{cases} k_{ij} = 1 \text{ if individual contains modality } j \\ k_{ij} = 0 \text{ else} \end{cases}$$

Table 1. Matrix of the TCC table data.

Table 2: Matrix of the CDT table data.

| TCC = | 1 | ... | j | ... | P |
|---|---|---|---|---|---|
| 1 | | | ⋮ | | |
| ⋮ | | | ⋮ | | |
| i | ... | ... | $x_{ij}$ | ... | ... |
| ⋮ | | | ⋮ | | |
| n | | | ⋮ | | |

| CDT= | 1 | ... | J | ... | $k_j$ |
|---|---|---|---|---|---|
| 1 | | | ⋮ | | |
| ⋮ | | | ⋮ | | |
| i | ... | ... | $k_{ij}$ | ... | ... |
| ⋮ | | | ⋮ | | |
| n | | | ⋮ | | |

In our approach the formula which will be used to enable this evaluation corresponds to the Khi² formula and is given by:

$$d^2(s,s) = \sum_{i=1}^{n} \frac{1}{f_{i.}} \left(\frac{k_{is}}{n_s} - \frac{k_{is}}{n_s}\right)^2 = n \sum_{i=1}^{n} \left(\frac{n_{is}}{n_s} - \frac{n_{is}}{n_s}\right)^2$$

## 4. Proposed Approach for Data Warehouse Reduction.

In our approach, we propose a method to reduce the number of dimensions p of a DW using the AGs and the MCA. We try to identify a subset p' of dimensions where p'<p, in which we can find the closest elements to the reference. Each individual returned by the GA represents a subset of dimensions and is considered as a possible solution. Every solution is represented by a chromosome containing as many genes as the DW dimensions. Each gene has two possible values: 1 if we keep the corresponding dimension in the solution and 0 otherwise. We then calculate for each solution, the distance between the reference and each fact contained in the extracted sample from the DW. We then consider the minimum value of the distances computed: if we choose to make the sum of all the distances computed we wouldn't be able to identify the facts with minimum values and that are close to the reference. The approach is based on the measure of similarity between two individuals.

The profile of interesting facts can then be ignored. Selecting the minimum in each step of the calculation helps identifying the facts that are the closest to the reference according to the distances measured with associated reductions. We have then two results at the same time: closest elements to the reference in the sample and associated reduction. In all cases, the lower the distance is, the most appropriate is the subset. The fitness function used to evaluate the quality of the individuals is the Khi² formula deduced from the MCA and presented above. This formula, applied to the multidimensional concept then becomes as follows :

$$d^2(i,i') = \frac{1}{p}\sum_{\mu=1}^{\alpha} \frac{(x_i - x_{i'})^2}{m/n} = \frac{n}{p}\sum_{\mu=1}^{\alpha} \frac{(x_i - x_{i'})^2}{m}$$

$p$ represents the number of dimensions of the DW, $n$ the number of elements in the sample, $\mu$ a modality or member of a dimension, $m$ number of occurrence of $\mu$, $\alpha$ number of modalities in dimensions. If the number of occurrences $m$ increases the modality is more frequent in the population.

The different steps of calculation and processing data are as follows:
1. Creating $P_{init}$ of $N$ individuals by the process of GAs corresponding to the first generation.
2. Selecting a sample of $N'$ elements from the DW to use for the calculation of distances.
3. Creation of the TCC table.
4. Creation of the CDT table.
5. Calculation of the distances.
6. Generation of the next generation with the operators cross over and mutation.

The algorithm corresponding to such a process is given by:

**FUNCTION GEN_P   (N, P)**

*1. for* each $ind_{i,1\leq i\leq N}$ ***do***
*2.     for* each $gene_{j,1\leq j\leq P}$ ***do***
*3.         $ind_i[gene_j] \leftarrow rondom(1,0)$*
*4.     end for*
*5. end for*

**END GEN_P$_{init}$**

**FUNCTION GEN_TCC (P, α)**

*6. for* each $fact_{i,1\leq i\leq N'}$ *do*
*7. $Tab_{i,j;1\leq i\leq N';1\leq j\leq P} \leftarrow fact_{i,j;1\leq i\leq N';1\leq j\leq P}$*
*8. end for*
*9. for* each $TAB_{i,j;1\leq i\leq N';1\leq j\leq P}$ *do*
*10. $TCC_{N',P} \leftarrow coding_{0,1}(TAB_{i,j}, \boldsymbol{\alpha}, P)$*
*11. end for*

**END GEN_TCC**

**FUNCTION GEN_TDC (P, α)**

*12. for* each $TCC_{N',P}$ *do*
*13. $TDC_{N',dim} \leftarrow coding_{0,1}(TCC_{N',P}, \boldsymbol{\alpha}, P)$*   (dim is the total number of dimension)
*14. end for*

**END GEN_TDC**

**FUNCTION COMPUTE (N, α)**

*15. $occ \leftarrow 0$*
*16. $som \leftarrow 0$*

**17.** *for each* $TDC_{i,j; 1 \leq i \leq N', 1 \leq j \leq dim}$ *do*
**18.** *if* $TDC_{i,j} = 1$ *then* $occ \leftarrow occ+1$
**19.** *end if*
**20.** *end for*
**21.** *for each* $ind_i[gene_j]$
**22.** *if* $ind_i[gene_j] = 1$ *then*
**23.** *for each* $TDC_{k,l; 1 \leq k \leq N', 1 \leq l \leq dim}$ *do*
**24.**     $A = math.pow((TDC_{k,l} - ref)), 2)/occ$
**25.**     $som \leftarrow som + A$
**26.** *end for*
**27.** *else*
**28.** *for each* $TDC_{k,l; 1 \leq k \leq N', 1 \leq l \leq dim\{j\}}$ *do*
**29.**     $A = math.pow((TDC_{k,l} - ref)), 2)/occ$
**30.**     $som \leftarrow som + A$
**31.** $dist[\,] \leftarrow \min(som)$
**32.** *end for*

**END COMPUTE**

Selection operator, we only keep distances computed and corresponding to individuals that are lower than a certain limit

***FUNCTION SELECTION (N, limit)***

**33.** *for each* $ind_i[gene_j] \in G_i$ *do*
**34.** *if* $ind_i(dist) < limit$
**35.** *then* $ind_i[gene_j] \in G_{i+1}$
**36.** *end if*
**37.** *end for*

**END SELECTION**

Crossing of two individuals to generate two children (crossover operator)

***FUNCTION CROSSOVER ()***

**38.** $cross\_pos \leftarrow random(1,2,3,4,5)$
**39.** $aux[k \leftarrow 0]$
**40.** *for each* $ind_i[gene_j]$ *do*
**41.** *for each* $ind_i[gene_{cross\_pos}]$ *do*
**42.** $aux[\,][k] \leftarrow ind_i[gene_j]$
**43.** $ind_i[gene_j] \leftarrow ind_{i+1}[gene_j]$
**44.** $ind_i[gene_j] \leftarrow aux[k][\,]$
**45.** *end for*
**46.** *end for*

**END CROSSOVER**

/* Mutation of two children with a probability $p_m$ (mutation operator)   */
***FUNCTION MUTATION ()***

**47.** $mut\_pos \leftarrow random(1,2,3,4,5,6)$
**48.** $nb\_mut \in \{1,2,3,4,5,6\}$
**49.** *for k from 1 to nb_mut do*
**50.** *for each* $ind_i[gene_j]$ *do*

**51.** if $ind_i[gene_{mut\_pos}] == 0$ then $ind_i[gene_{mut\_pos}] \leftarrow 1$ else $ind_i[gene_{mut\_pos}] \leftarrow 0$
**52.** end if
**53.** end for
**54.** end for
<p align="center">**END MUTATION**</p>

## 5. Application on an Example.

In this section, it is proposed to apply the approach and the algorithm presented in section 4 on the example of DW shown in Figure1. The objective of the DW is to collect information on travellers at the border state offices posts to constitute a Big Data tables to learn about their movements and monitor their traffic. The DW is formed by six DIMENSIONS TABLES and a central FACTS TABLE.

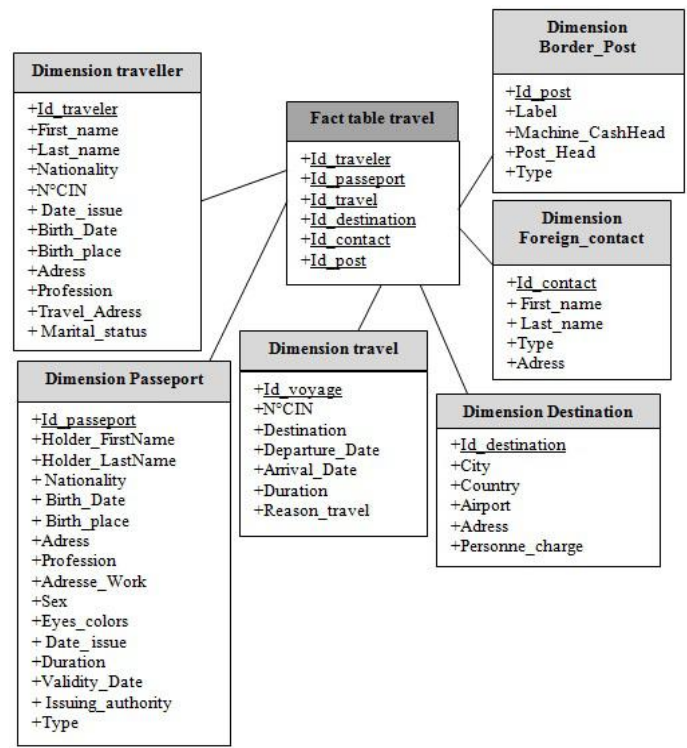

Figure 1: Fact table and dimensions of the DW.

The different dimensions are {TRAVELLER, PASSEPORT, TRAVEL, DESTINATION, CONTACT, BORDER CONTROL}. The DW contains qualitative data where each dimension has a limited and fixed number of possible values presented as follows:
TRAVELLER{ Foreign, Citizen}
PASSEPORT{ Normal, Diplomatic, Mission }
TRAVEL{ Leisure, Professional }
DESTINATION{Asia, Africa, Western Europe, South America, Australia, North America, Eastern Europe  }
CONTACT{Person, Company}
BORDER CONTROL{Aerial, Earth, Maritime}

As we have explained in section 3.2, two special structures are considered in the MCA theory during the process of calculation: the TCC and the CDT. In the following paragraph, we present these two structures and their proposed adaptation to our example.

The MCA is used to measure the quality of the solutions returned by the process of GAs by applying the Khi² formula. We consider a sample of N' facts extracted from the DW since the Khi² formula contains the parameter m representing the occurrence number of each modality in the sample. The TCC represents the initial values taken by the dimensions of that sample but in binary notation. The following Table 3 presents the sample with discrete values we propose to consider:

Table 3: Extracted sample of facts from the DW.

|  | *DIM1 TRAVELLER* | *DIM2 PASSEPORT* | *DIM3 TRAVEL* | *DIM4 DESTINATION* | *DIM5 CONTACT* | *DIM6 BORDER CONTROL* |
|---|---|---|---|---|---|---|
| *IND1* | Foreign | Normal | Leisure | Asia | Person | Aerial |
| *IND2* | Citizen | Diplomatic | Leisure | Africa | Company | Earth |
| *IND3* | Citizen | Mission | Professional | Western Europe | Person | Maritime |
| *IND4* | Foreign | Diplomatic | Professional | South America | Person | Maritime |
| *IND5* | Citizen | Mission | Leisure | Asia | Company | Maritime |
| *IND6* | Foreign | Normal | Professional | Africa | Person | Earth |
| *IND7* | Citizen | Diplomatic | Leisure | Australia | Company | Aerial |
| *IND8* | Foreign | Normal | Leisure | South America | Person | Maritime |
| *IND9* | Foreign | Diplomatic | Professional | North America | Company | Earth |
| *IND10* | Citizen | Mission | Professional | Eastern Europe | Person | Earth |

For each sample extracted, a TCC, corresponding to the dimensions of the DW, will be generated and will represent the same information initially stored in the data warehouse but encoded in binary notation. Each value taken by the fact in the corresponding dimension will be encoded in a binary notation. The corresponding TCC to the sample then considered is given by the following Table 4:

Table 4: TCC corresponding to the extracted sample.

|  | *DIM1* | *DIM2* | *DIM3* | *DIM4* | *DIM5* | *DIM6* |
|---|---|---|---|---|---|---|
| *IND1* | 01 | 010 | 01 | 000001 | 10 | 001 |
| *IND2* | 10 | 001 | 10 | 001000 | 01 | 010 |
| *IND3* | 10 | 100 | 10 | 010000 | 10 | 100 |
| *IND4* | 01 | 001 | 10 | 100000 | 10 | 100 |
| *IND5* | 10 | 100 | 01 | 000001 | 01 | 100 |
| *IND6* | 01 | 010 | 10 | 001000 | 10 | 010 |
| *IND7* | 10 | 001 | 01 | 000010 | 01 | 001 |
| *IND8* | 01 | 010 | 01 | 100000 | 10 | 100 |
| *IND9* | 01 | 001 | 10 | 010000 | 01 | 010 |
| *IND10* | 10 | 100 | 10 | 001000 | 10 | 010 |

The second structure presented in the MCA theory concerns the CDT that represents the whole modalities of the DW and not only the dimensions. The TCC represents a collection of binary values, 0 or 1 arranged according to the dimensions of the DW. This table couldn't then give an understandable idea about the different modalities of the DW. Consequently, the CDT will be used to give a more detailed view of the data in the DW according to the modalities considered.

This table will be used the Khi² formula during the calculation process. It also helps identifying some necessary parameters for the computation such as the parameter m representing the number of occurrences of each modality. The corresponding CDT to the sample considered previously is presented by the following Table5:

Table 5: CDT corresponding to the extracted sample.

| | dim1 | | dim2 | | | dim3 | | dim4 | | | | | | | dim5 | | dim6 | | | |
|---|---|---|---|---|---|---|---|---|---|---|---|---|---|---|---|---|---|---|---|---|
| | Mod1 | Mod2 | Mod1 | Mod2 | Mod3 | Mod1 | Mod2 | Mod1 | Mod2 | Mod3 | Mod4 | Mod5 | Mod6 | Mod7 | Mod1 | Mod2 | Mod1 | Mod2 | Mod3 | |
| | 0 | 1 | 0 | 1 | 0 | 0 | 1 | 0 | 0 | 0 | 0 | 0 | 0 | 1 | 1 | 0 | 0 | 0 | 1 | IND1 |
| | 0 | 1 | 0 | 0 | 1 | 0 | 1 | 0 | 0 | 1 | 0 | 0 | 0 | 0 | 0 | 1 | 0 | 1 | 0 | IND2 |
| | 1 | 0 | 0 | 1 | 0 | 1 | 0 | 0 | 1 | 0 | 0 | 0 | 0 | 0 | 0 | 1 | 1 | 0 | 0 | IND3 |
| | 0 | 1 | 1 | 0 | 0 | 1 | 0 | 1 | 0 | 0 | 0 | 0 | 0 | 0 | 1 | 0 | 1 | 0 | 0 | IND4 |
| CDT= | 0 | 1 | 0 | 0 | 1 | 0 | 1 | 0 | 0 | 0 | 0 | 1 | 0 | 0 | 0 | 1 | 1 | 0 | 0 | IND5 |
| | 1 | 0 | 0 | 1 | 0 | 1 | 0 | 0 | 0 | 0 | 1 | 0 | 0 | 0 | 0 | 1 | 0 | 1 | 0 | IND6 |
| | 0 | 1 | 1 | 0 | 0 | 1 | 0 | 0 | 0 | 0 | 0 | 1 | 0 | 1 | 0 | 0 | 0 | 0 | 1 | IND7 |
| | 1 | 0 | 0 | 1 | 0 | 0 | 1 | 0 | 0 | 0 | 0 | 0 | 1 | 0 | 1 | 0 | 1 | 0 | 0 | IND8 |
| | 0 | 1 | 1 | 0 | 0 | 1 | 0 | 0 | 1 | 0 | 0 | 0 | 0 | 0 | 1 | 0 | 0 | 1 | 0 | IND9 |
| | 0 | 1 | 0 | 0 | 1 | 1 | 0 | 0 | 0 | 1 | 0 | 0 | 0 | 0 | 0 | 1 | 0 | 1 | 0 | IND10 |
| **ref** | **1** | **0** | **0** | **1** | **0** | **0** | **1** | **0** | **0** | **0** | **1** | **0** | **0** | **0** | **0** | **1** | **0** | **0** | **1** | |

The CDT and the TCC obtained transform the discrete qualitative data into a binary set of data which facilitates their calculation using the Khi² adapted formula. This formula is used to measure the degree of similarity between individuals. Each individual will have an encoding of 19-bit representing its whole characteristic profile. The encoding of the individuals in this table is not aleatory; it should respect the following findings to preserve the integrity of the table: the number of bits encoded 1 in a line is equal to P (number of dimensions) and the number of occurrences m for each dimension is equal to N' (number of elements in the sample)

In the process of GAs, each solution will be encoded by a 6 bits chromosome which corresponds to the number of dimensions. Each gene is coded 1 if the corresponding dimension exists for the individual and 0 otherwise. The total number of possible solutions is then $2^6 = 64$.

## 6. Results Interpretation.

As already defined, the purpose of the study is to reduce the DW dimensions by determining the most suitable subset of dimensions where the sample of facts retained contains the closest to the reference. For this, it is proposed to measure the distance between the reference profile $x_{ref}$ and all the other individuals $x_i$ of the selected sample taking into consideration the reduced dimensions. We will then try to identify the elements in the sample which profile is similar to the reference profile.

In the calculation process, it is considered to compute ten values of distances. It is then proposed to retain the minimum one as the distance between the reference and that element. If we sum all the measures, we couldn't then identify the closest individuals to the reference since they will be hidden by the sum. For each step of computation, we should take into consideration the solutions returned by the process of the GAs.

This calculation could be performed in two ways: the first one is to compute the distances and keep in the CDT the dimensions retained by the GA and replace all the rest of the values by 0. The second one consists on totally removing the not retained dimensions in the GAs proposal and excluding them from the calculation as well as the reference. We then consider the minimum value among all the measures.

An illustrative example is given as follows: The solution returned by the process of the GA is: 011010 where we choose to keep the subset of dimensions (235). The distances were measured between the reference given in the CDT and each element of the sample considered. We then obtain the results presented in the Table 6 below:

Table 6: Results corresponding to the reduction.

| INDIVIDUALS | $x_{ref}, x_1$ | $x_{ref}, x_2$ | $x_{ref}, x_3$ | $x_{ref}, x_4$ | $x_{ref}, x_5$ | $x_{ref}, x_6$ | $x_{ref}, x_7$ | $x_{ref}, x_8$ | $x_{ref}, x_9$ | $x_{ref}, x_{10}$ |
|---|---|---|---|---|---|---|---|---|---|---|
| $D^2_{Khi}(x_{ref}, x_i)$ | 3,573 | 3,867 | 3,573 | 5,147 | 3,867 | 3,573 | 4,507 | 3,573 | 5,147 | 4,507 |

The minimum distances were obtained for individuals $x_1, x_3, x_6$ and $x_8$ which explains that with the reduction (235), we first detect 4 minimum values which mean that the sample of ten elements considered contains four individuals that are the closest to the reference with an average of 40%. The value that will be retained in the processing is $D^2_{Khi} = 3.57$ as the minimum value of distance corresponding to this reduction.

The same previous reasoning was also applied for all the other reductions, retaining for each time the minimum value computed. We then obtain the following Figure 2:

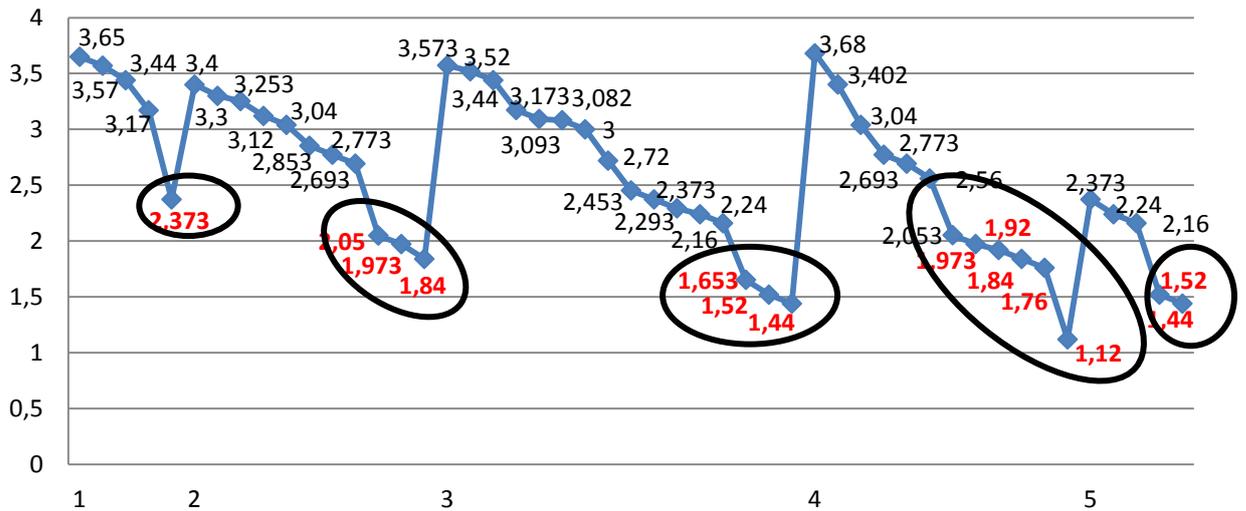

Figure 1: Distance measurement with reference with reduction

According to the previous curve, for each number of retained dimensions, we identify two subsets: one for minimum distances and the other for maximum ones, which means that for each group of solutions we can identify maximum and minimum measures. In our approach it is proposed to conserve only groups corresponding to the minimum results to get closer as much as possible to the reference. Providing the analyst with these results can assist him in choosing the right reduction configuration.

Distance computation is mainly used for segmentation process to identify similar groups containing individuals with similar characteristics. Our approach is based on a measure of similarity between a reference and all the individuals of the sample considered. We can by the way change the configuration of the reference whenever the analyst wishes.

## 7. Conclusions and Perspectives.

Collecting and storing information in a DW creates huge data structures which analysis is a complex task. These DW could be visualized in multidimensional structures called OLAP cubes according to multiple analysis axes. To make the analysis process of the DW easier, data reduction methods have been proposed where the reduction may concern either the number of facts (lines) or dimensions (columns) of the warehouse. Our approach is based on the use of GAs and the MCA, to express the similarity between a reference and a group of individuals to find the most similar individual in the sample to the reference by using the Khi² formula. The method provided is used only for qualitative data which is a great limitation because analysis can also concern quantitative data. The process identifies groups of distances corresponding to different distances some are maximum and others are minimum. The analyst can then choose which solution fits the requirements of the analysis. In all cases, the reduction corresponds to a loss of information either by replacing the values by 0 for the not retained dimensions or by eliminating them from the computation. In our case, we have applied a technique based on distances measurement widely used for segmentation and clustering to identify the most appropriate dimensions reductions in a DW.